\documentclass{article}
\usepackage{ijcai23}

\usepackage{times}
\usepackage{soul}
\usepackage{url}
\usepackage[hidelinks]{hyperref}
\usepackage[utf8]{inputenc}
\usepackage[small]{caption}
\usepackage{graphicx}
\usepackage{amsmath}
\usepackage{amsthm}
\usepackage{booktabs}
\usepackage{algorithm}
\usepackage{algorithmic}
\usepackage{subfigure}
\usepackage[switch]{lineno}

\usepackage{multirow}
\usepackage{pythonhighlight}
\usepackage{amssymb}
\usepackage{tabularray}

\title{\textsc{FLGo}: A Fully Customizable Federated Learning Platform}

\author{
Zheng Wang$^{1,2}$
\and
Xiaoliang Fan$^{1,2*}$\and
Zhaopeng Peng$^{1,2}$\and
Xueheng Li$^{1,2}$\and
Ziqi Yang$^{1,2}$\and\\
Mingkuan Feng$^{1,2}$\and
Zhicheng Yang$^{1,2}$\and
Xiao Liu$^3$\and
Cheng Wang$^{1,2}$
\affiliations
$^1$ Fujian Key Laboratory of Sensing and Computing for Smart Cities, School of Informatics, Xiamen University, Xiamen, China\\
$^2$ Key Laboratory of Multimedia Trusted Perception and Efficient Computing, Ministry of Education of China, Xiamen University, Xiamen, China\\
$^3$ School of Information Technology, Deakin University, Geelong, Australia\\
\emails
zwang@stu.xmu.edu.cn,
fanxiaoliang@xmu.edu.cn,
pengzhaopeng@stu.xmu.edu.cn
22920202202797@stu.xmu.edu.cn, lafinhana@outlook.com, 22920202202770@stu.xmu.edu.cn, 
zcyang@stu.xmu.edu.cn,
xiao.liu@deakin.edu.au, cwang@xmu.edu.cn
}

\begin{document}
\maketitle

\begin{abstract}
    Federated learning (FL) has found numerous applications in healthcare, finance, and IoT scenarios.
    Many existing FL frameworks offer a range of benchmarks to evaluate the performance of FL under realistic conditions. However, the process of customizing simulations to accommodate application-specific settings, data heterogeneity, and system heterogeneity typically remains unnecessarily complicated. This creates significant hurdles for traditional ML researchers in exploring the usage of FL, while also compromising the shareability of codes across FL frameworks. To address this issue, we propose a novel lightweight FL platform called FLGo, to facilitate cross-application FL studies with a high degree of shareability. Our platform offers 40+ benchmarks, 20+ algorithms, and 2 system simulators as out-of-the-box plugins. We also provide user-friendly APIs for quickly customizing new plugins that can be readily shared and reused for improved reproducibility. Finally, we develop a range of experimental tools, including parallel acceleration, experiment tracker and analyzer, and parameters auto-tuning. FLGo is maintained at  \url{flgo-xmu.github.io}. 
\end{abstract}

\section{Introduction}
\begin{figure}
\centering
\includegraphics[scale=0.5]{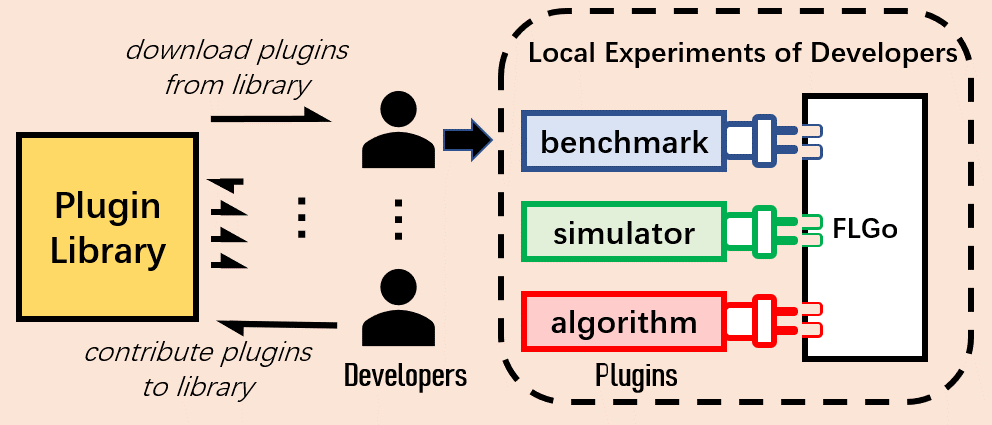}
\caption{FLGo implements benchmarks, simulators, and algorithms as independent plugins, where developers can download plugins and share locally customized plugins through a plugin library.}
\label{fig1}
\end{figure}

\begin{table*}
\centering
\begin{tabular}{c|c|c|c|c|c|c|c|c}
                                    & \textbf{TFF} &  \textbf{LEAF} &\textbf{PySyft} & \textbf{FedML} & \textbf{FederatedScope} & \textbf{FedScale} & \textbf{Flower} & \textbf{FLGo}  \\ 
\hline
\textbf{Data Heterogeneity}         & 1            & 1               & 1             & 1              & 1                       & 1                 & 1               & 1              \\
\textbf{System Heterogeneity}       & 0           &    $\times$          &   $\times$           & 0              & 1                       & 1                 & 0               & 1              \\
\textbf{High-Level API}             & 0             &    $\times$       &  0            & 1              & 1                       & 1                 & 1               & 1              \\
\textbf{Multi-Architecture}     &       $\times$       &         $\times$     &       $\times$        & 1              & 1                       & 0                 & 0               & 1              \\
\textbf{Asynchrounous}      &        $\times$      &          $\times$       &    $\times$           &    0            & 1                       &      $\times$            &   $\times$              & 1              \\
\textbf{Experiment Manager}       &       $\times$       &      $\times$           &     $\times$          & 1              & 1                       & 0                  &        $\times$         & 1        \\ 
\textbf{Behavior Customization} &        $\times$      & $\times$                &      $\times$         &  1  & 1                      & 0                 & 0               & 1              \\
\textbf{Benchmark Customization}  &        $\times$      &            $\times$     &        $\times$       & 0              &        $\times$                 &           $\times$       &     $\times$            & 1              \\
\textbf{Heterogeneity Customization}         &          $\times$    &      $\times$     &     $\times$      & 0           &   0    & 0   & 0    & 1 \\   
\end{tabular}
\caption{Comparison on FL Frameworks. 0 represents partially implemented, 1 represents fully implemented, $\times$ represents not implemented.}
\label{tab_rw}
\end{table*}
Federated learning (FL) has attracted wide attention from both academics and industries \cite{advance,he2020fedml,li2020federated}. Its characteristic of privacy protection has made it a popular choice for data security compliance applications, including medicine \cite{liu2021feddg}, finance \cite{long2021banking}, internet of things \cite{iot}, etc. 
Researchers have proposed numerous FL algorithms to address both data \cite{wang2021fedfv,li2020federated,karimireddy2020scaffold} and system heterogeneity\cite{li2020federated,wang2020tackling,wang2022fedgs}. However, the effectiveness of these algorithms is limited \cite{li2022niidbench}, where each algorithm can only bring non-trivial improvement, or even be worse than FedAvg in specific contexts \cite{li2022niidbench}. Therefore, it is essential for engineers to conduct quantitative evaluations on available FL algorithms before deploying them. 

 Recently, several platforms were proposed to enable researchers to fairly evaluate FL algorithms by providing multiple benchmarks. FedScale\cite{lai2022fedscale} uses systematic datasets to simulate realistic mobile-device settings and real-world federated datasets, while FederatedScope\cite{xie2023federatedscope} employs an architecture driven by events to customize the behaviors of different participants. 
 However, the complexity involved in implementing customized settings, such as different applications, heterogeneous data, and system heterogeneity, is commonly overlooked. This substantial limitation poses a significant barrier for machine learning researchers seeking to employ FL frameworks in practical scenarios.
 Additionally, the high complexity of customization further harms the code shareability due to the absence of standardized guidelines for customization, even when employing the same FL framework.  
 In short, addressing these issues is crucial in order to develop an adaptable FL framework that  facilitates cross-application FL studies with a high degree of shareability.

To tackle these problems, we develop FLGo, a novel lightweight FL platform that streamlines cross-application FL research with a high degree of shareability. By leveraging the \textit{plugin library}, developers can easily distribute their customization as \textit{plugins}, as illustrated in Fig.\ref{fig1}. 
Our main contributions are summerized as follows:

\begin{itemize}
    \item We propose a novel FL platform, FLGo, to support customization with high shareability, where benchmarks, algorithms and simulators are implemented as plugins enable the fast development of cross-application FL scenarios.
    \item We provide 40+ comprehensive FL benchmarks, 20+ algorithms, and 2 system simulators in FLGo. We also develop useful tools to support various experimental purposes including parallel acceleration, experiment tracker and analyzer, and parameters auto-tuning.
    \item We conduct comprehensive experiments on 6 datasets to evaluate the functionality of FLGo platform by customizing various data and system heterogeneity settings.
\end{itemize}

\section{Related Works}
\paragraph{Existing frameworks for FL research.}    
Extensive federated frameworks were developed to better support FL studies in both academic and industrial scenarios \cite{he2020fedml,liu2021fate}. LEAF \cite{caldas2018leaf} provides several open-source federated benchmarks. TFF \cite{bonawitz2019towards} focuses on the application of large-scale mobile devices and PySyft \cite{ryffel2018generic} focuses on deep learning for data security. FedML \cite{he2020fedml} facilitates equitable comparison of algorithms through the provision of benchmarks, multiple topologies, and diverse computing paradigms. FedScale \cite{lai2022fedscale} provides real-world datasets to evaluate the statistical efficiency and system efficiency and improve the efficiency under large scalability. FederatedScope \cite{xie2023federatedscope} employs an event-driven architecture to enable arbitrary customization on parties' behaviors. Flower \cite{beutel2020flower} supports different backends and is also scalable. However, the challenge of evaluating FL algorithms for customized applications, data heterogeneity, and system heterogeneity is often overlooked. This issue presents a natural divide between conventional machine learning developers and the FL community. By addressing this challenge and enabling greater compatibility and shareability, FL can become more accessible to a wider range of developers, further promoting its adoption and advancement.

\paragraph{Existing frameworks are inadequate.} Although great efforts have been made in developing FL community, these frameworks failed to 1) fast convert any ML task into a federated benchmark to attract more researchers in the traditional ML community, and 2) provide high-level APIs for customization on heterogeneity. To simulate more realistic and complex scenarios when adopting FL, we propose FLGo. We make a comparison between FLGo and other frameworks to distinguish the advantages of FLGo in Table.1.
\section{FLGo Features}
\subsection{Overview}
\begin{figure}[tbp]
\centering
\includegraphics[scale=0.55]{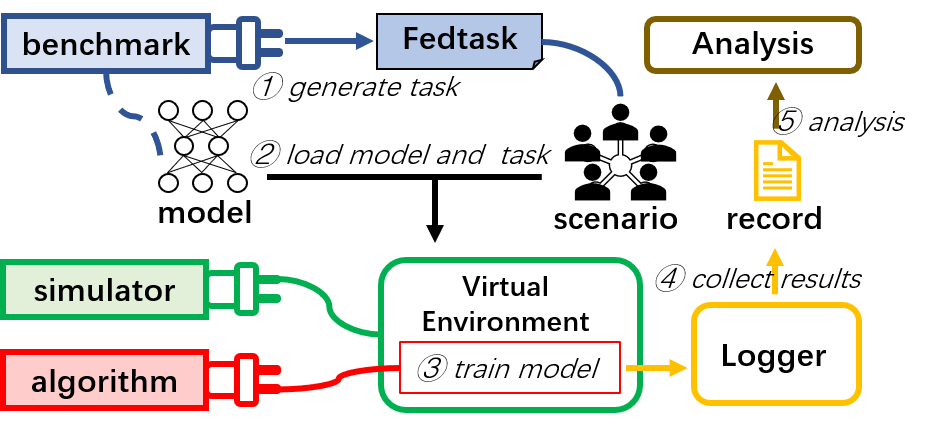}
\caption{The main workflow in FLGo.}
\label{fig2}
\end{figure}

\begin{table*}
\centering
\begin{tabular}{l|l|l|l} 
\hline\hline
\textbf{Category}               & \textbf{Task}           & \textbf{Scenario}    & \textbf{Datasets}                                                                                                 \\ 
\hline
\multirow{3}{*}{\textbf{CV}}    & Classification          & Horizontal \& Vertical & \begin{tabular}[c]{@{}l@{}}CIFAR10, CIFAR100, MNIST, FashionMNIST, \\FEMNIST, EMNIST, SVHN\end{tabular}  \\
                                & Detection               & Horizontal           & Coco, VOC                                                                                                         \\
                                & Segmentation            & Horizontal           & Coco, SBDataset                                                                                                      \\ 
\hline
\multirow{3}{*}{\textbf{NLP}}   & Classification          & Horizontal           & Sentiment140, AG\_NEWS, sst2                                                                                      \\
                                & Text Prediction         & Horizontal           & Shakespeare, Reddit                                                                                               \\
                                & Translation             & Horizontal           & Multi30k                                                                                                          \\ 
\hline
\multirow{3}{*}{\textbf{Graph}} & Node Classification     & Horizontal           & Cora, Citeseer, Pubmed                                                                                            \\
                                & Link Prediction         & Horizontal           & Cora, Citeseer, Pubmed                                                                                            \\
                                & Graph Classification    & Horizontal           & Enzymes, Mutag                                                                                                    \\ 
\hline
\textbf{Rec}                    & Rating Prediction       & Horizontal\&Vertical & Ciao, Movielens, Epinions, Filmtrust, Douban                                                                      \\ 
\hline
\textbf{Series}                 & Time series forecasting & Horizontal           & Electricity, Exchange Rate                                                                                        \\ 
\hline
\textbf{Tabular}                & Classification          & Horizontal           & Adult, Bank Marketing, Heart Disease                                                                                             \\ 
\hline
\textbf{Synthetic}              & Regression              & Horizontal           & Synthetic, DistributedQP                                                                                 \\
\hline
\end{tabular}
\caption{An overview of benchmarks in FLGo.}
\label{tab_bmk}
\end{table*}

\begin{table*}
\centering
\begin{tabular}{c|c|c|c|c|c} 
\hline
\textbf{Name}                      & \textbf{Feature Skew} & \textbf{Label Skew} & \textbf{Concept Drift} & \textbf{Concept Shift} & \textbf{Quantity Skew}  \\ 
\hline
IIDPartitioner \shortcite{mcmahan2017communication}                  &                       &                     &                        &                        & $\checkmark$            \\
DiversityPartitioner \shortcite{mcmahan2017communication}            & $\checkmark$          & $\checkmark$        &                        &                        &                         \\
DirichletPartitioner \shortcite{ming2019measure}            & $\checkmark$          & $\checkmark$        &                        &                        & $\checkmark$            \\
GaussianPerturbationPartitioner \shortcite{li2022niidbench} & $\checkmark$          &                     &                        &                        & $\checkmark$            \\
IDPartitioner               & $\checkmark$          & $\checkmark$        & $\checkmark$           & $\checkmark$           & $\checkmark$            \\
VerticalPartitioner      & $\checkmark$          &                     &                        &                        & $\checkmark$            \\
NodeLouvainPartitioner \shortcite{fedsage}          & $\checkmark$          &                     &                        &                        &                         \\
\hline
\end{tabular}
\caption{Partitioners for data heterogeneity in FLGo.}
\label{tab_noniid}
\end{table*}
The workflow in FLGo framework is as shown in Fig.\ref{fig2}. First, the blue \textit{benchmark} plugin generates a static federated task on the disk, which describes how the data is distributed among participants. Second, FLGo system will load the model and the stored task to construct the simulation. Then, the red \textit{algorithm} plugin will optimize the model under the virtual environment created by the green \textit{simulator} plugin. Finally, the \textit{logger} will trace the running-time variables and save them as a record on the disk for further analysis. 

\subsection{Benchmark Module}

\subsubsection{Comprehensive Benchmarks}
We have provided 40+ out-of-the-box FL benchmarks in Table.\ref{tab_bmk}. The details of each dataset are on the website\footnote{https://flgo-xmu.github.io}. These benchmarks enable FLGo to support a wide range of studies through three characteristics:
\begin{itemize}
    \item \textbf{Cross-Category.} We implement the templates of benchmark plugins for different types of input data including images, text, graphs, series, and tables.
    \item \textbf{Cross-Application.} Tasks of different applications like finance, healthcare, and IoT scenarios are integrated.
    \item \textbf{Cross-Architecture.} Benchmark plugins can also support FL with different architectures like horizontal FL, vertical FL, decentralized FL, and hierarchical FL.
\end{itemize}

\subsubsection{Customization}
\begin{python}\label{code1}
/*Code.1
import flgo
flgo.gen_benchmark(name, config, target_path, 
data_type, task_type)
\end{python}
We design a general paradigm to ease the customization of various benchmarks. In addition, we provide templates of benchmark plugins for different categories of input data and different types of tasks. By using the templates, developers can integrate their customized benchmarks with only one line of code as shown in Code.1.

 \begin{figure*}
\centering
\includegraphics[scale=0.5]{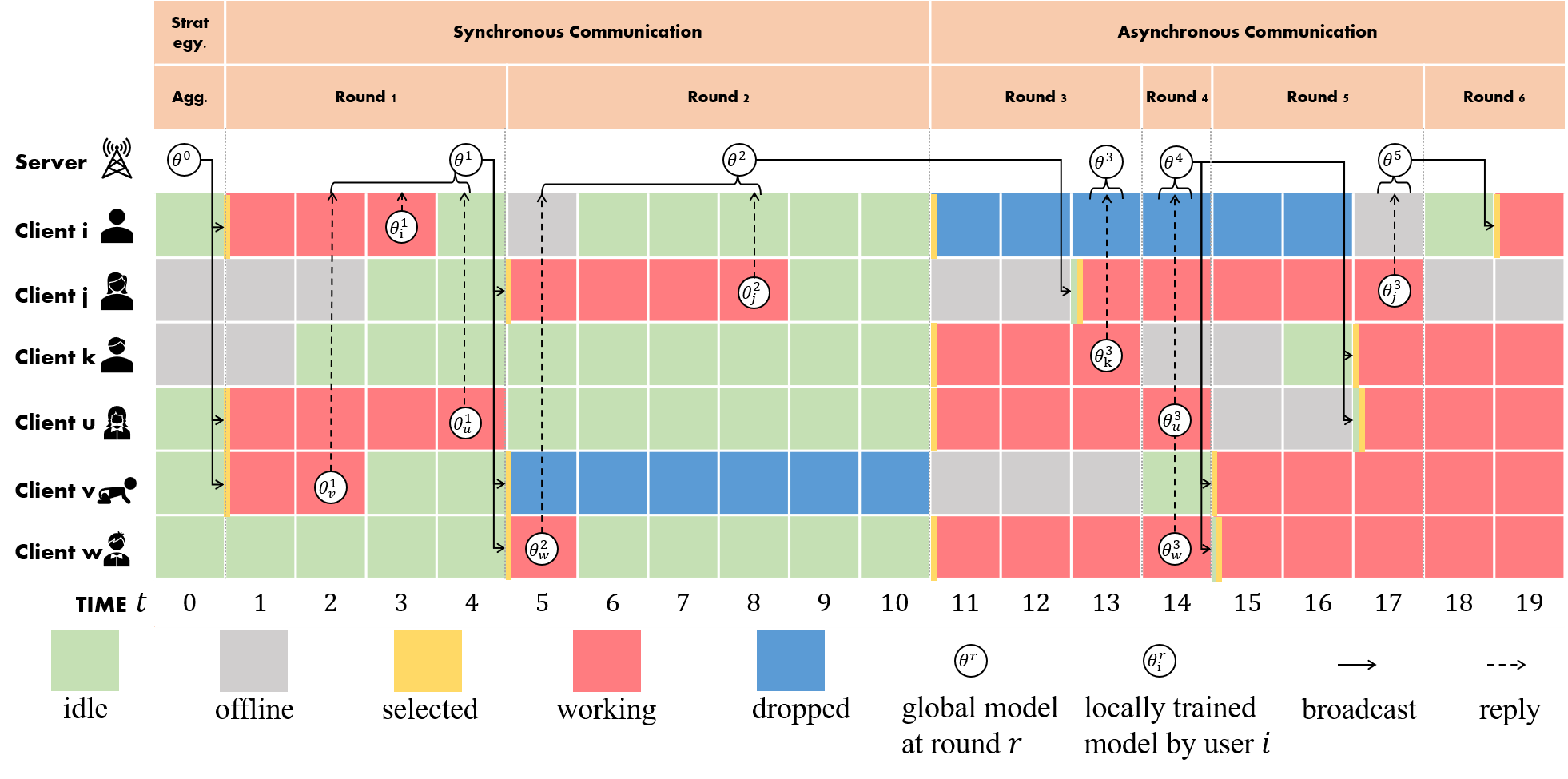}
\caption{FLGo simulates synchronous and asynchronous scenarios by using the global clock and the client-state machine, where each client's state can shift either as time passes or when a particular condition is reached (e.g., being selected by the center).}
\label{fig3}
\end{figure*}

\subsubsection{Paritioner for Data Heterogeneity}
 To support studies in FL with different types of data heterogeneity categorized by \cite{advance}, we create reusable data partitioners to simulate real-world local data distributions. In this way, developers can arbitrarily combine the benchmark plugins and partitioners together to design the data heterogeneity. We also provide APIs for customizing new partitioners. We list the data heterogeneity currently supported by FLGo in Table.3. 

\subsection{Simulator Module}
Existing frameworks \cite{xie2023federatedscope,lai2022fedscale} provided realizations of particular system heterogeneity as benchmarks, but they fail to support APIs for customization on the system heterogeneity.
Since both different types and degrees of system heterogeneity have non-trivial impacts on the training effectiveness \cite{lai2022fedscale,wang2022fedgs}, it's essential to allow developers to customize the system heterogeneity. To this end, we first conclude four types of system heterogeneity that were investigated in previous FL studies:
\begin{itemize}
    \item \textbf{Availability.} If a client is unavailable, the server cannot select it to join the model training \cite{wang2022fedgs}.
    \item \textbf{Responsiveness.} Responsiveness describes the length of the period for the server to wait for the responses of participants \cite{tifl}.
    \item \textbf{Completeness.} The client model updates may be incomplete\cite{li2020federated,wang2020tackling}.
    \item \textbf{Connectivity.} Some participants may lose connection for a long period \cite{jiang2023taming} even if they had been selected to join.
\end{itemize}

According to the aforementioned four types of system heterogeneity, we design 2 simulators, where one is based on synthetic data and another one is based on real-world data \cite{lai2022fedscale}. We first develop a client-state machine and a global virtual clock to simulate heterogeneous systems. For example in Fig.\ref{fig3}, it takes client $i$ 3 units of time to finish local training at the first round, and the length of time units that are spent by client $u$ is 4. Based on the global clock and the client-state machine, we then design easy-to-use APIs for developers to directly define how the state will shift for each participant at each moment or each round. In this way, the four types of system heterogeneity can be easily customized. 

\subsection{Algorithm Module}
In FLGo, each algorithm plugin is described by parties with their behaviors. We conclude the main features of the algorithm module of FLGo in the following subsections.
 \subsubsection{Time-based asynchronism}
 Most of the previous studies in asynchronous FL are based on round-wise asynchronism, which cannot be fairly compared with the synchronous FL methods since the time cost of each round is overlooked \cite{Xie2019AsynchronousFO,nguyen2022federated}. To this end, we implement the asynchronous FL algorithms based on the global virtual clock, where the developers should consider the behaviors of parties at each time unit when realizing asynchronous algorithms. For example, the server of fully asynchronous FL in Fig.\ref{fig3} samples participants once they become available (e.g. client $j$ at the 13th time unit) in round $3$, and receives the locally trained model from client $k$ at the same time unit. As a comparison, the synchronous FL server in rounds 1 and 2 will wait for the slowest participant at each iteration. In this way, the algorithms in syncFL and asyncFL are comparable under the same metric of time, which enables developers to search for the best suitable strategy under more realistic settings.

\subsubsection{Flexible Communication}

To enable different communication behaviors of parties, we follow the gRPC protocol to model the communication processes. Every time a party sends a message to a remote one, it will receive a response that is returned by a corresponding remote function. The transferred information is organized as key-value pairs for high flexibility. By registering the message handler to the specific message, developers can arbitrarily customize the communication process. For example in above Code.2, a party can send a specific message 'forward' to ask another remote party for partial activations by registering this message to the action of forward computing. 
\begin{python}\label{code2}
/*Code.2
import flgo.algorithm.vflbase

class ActiveParty(vflbase.ActiveParty):
    def iterate(self):
        ...
        # prepare the sending package
        pkg = self.set_message(\
              mtype='forward', package={})
        # request activations from the party
        acts = self.communicate_with(\
               party.id, pkg)['res']
        
class PassiveParty(vflbase.PassiveParty):
    def initialize(self):
        # register the action to 'forward'
        self.actions = {'forward':\ 
                       self.my_forward}

    # define the action
    def my_forward(self, package):
        ...
        return {'res':activations}
\end{python}
\subsubsection{Cross-Architecture}
To ease the development of algorithms with different architectures (i.e. horizontal FL \cite{mcmahan2017communication}, vertical FL \cite{Liu2022VerticalFL}, decentralized FL\cite{beltrán2023decentralized}, and hierarchical FL \cite{Liu2019ClientEdgeCloudHF}), we respectively conclude general paradigms for each type of architecture based on previous works. 
By using these paradigms, developers can easily customize their algorithms by only writing a few codes to replace the corresponding parts in the paradigm. 
\begin{figure}
\centering
    {%
        \includegraphics[width = .49\linewidth]{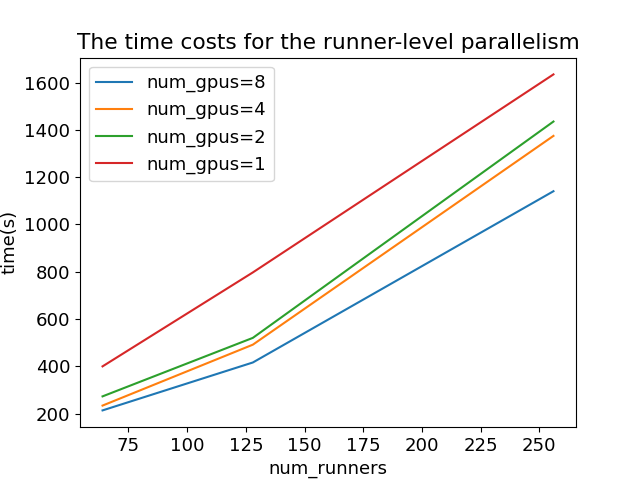}
        \includegraphics[width = .49\linewidth]{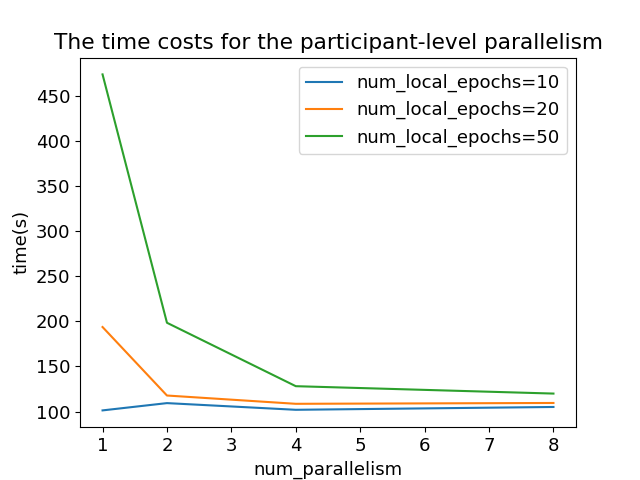}}

\caption{The acceleration by two levels of parallelism: (left) runner-level, and (right) participant-level. \label{fig5}}
\end{figure}
\subsection{Experiment Tools}

\begin{table*}
\centering
\begin{tabular}{|c|c|c|c|c|c|c|}
\hline
        & \multicolumn{3}{|c|}{\textbf{Data Heterogeneity}}& \multicolumn{2}{|c|}{\textbf{System Heterogeneity}}& \multicolumn{1}{|c|}{\textbf{Asynchronous}}\\
\hline
\textbf{Dataset} & Cifar 10                                                        & Cora                                                          & Shakespeare                                                         & Fashion Mnist                                                  & Reddit                                                         & SVHN\\
\hline
\textbf{Task}    & \begin{tabular}[c]{@{}c@{}}Image \\Classification~\end{tabular} & \begin{tabular}[c]{@{}c@{}}Node \\Classification\end{tabular} & \begin{tabular}[c]{@{}c@{}}Next-Character \\Prediction\end{tabular} & \begin{tabular}[c]{@{}c@{}}Image \\Classification\end{tabular} & \begin{tabular}[c]{@{}c@{}}Next-word \\Prediction\end{tabular} & \begin{tabular}[c]{@{}c@{}}Image \\Classification\end{tabular}\\
\hline
\textbf{Model}   & \begin{tabular}[c]{@{}c@{}}Two-layer \\CNN\end{tabular}         & \begin{tabular}[c]{@{}c@{}}Two-layer \\GCN\end{tabular}       & \begin{tabular}[c]{@{}c@{}}Stacked \\LSTM\end{tabular}              & \begin{tabular}[c]{@{}c@{}}Two-layer \\CNN\end{tabular}        & \begin{tabular}[c]{@{}c@{}}Stacked \\LSTM\end{tabular}         & \begin{tabular}[c]{@{}c@{}}Two-layer \\CNN\end{tabular} \\
\hline
\end{tabular}
\caption{Experimental datasets and corresponding models.}
\label{table4}
\end{table*}
\subsubsection{Parallel Acceleration}
We provide two-level parallels in FLGo to accelerate the FL training process in simulations. The first one is the runner-level, where each runner $r=(\mathcal{A}, \theta, G, \mathcal{T}, s)$ denotes using the algorithm $\mathcal{A}$ to optimize the global model $\theta$ with the hyperparameters $G$ on the federated task $\mathcal{T}$ under the environment created by the simulator $s$. The runner-level parallelism enables developers to use multiple GPUs to save efficiency, where devices will be  automatically scheduled by the device scheduler module to avoid out-of-memory errors. A runner queue is used to promise that all the runners will successfully finish their tasks, whereas the losers will be automatically put back in the queue again. The second-level parallelism is the participant-level, where each participant denotes a virtual client of each runner. The participant-level parallelism accelerates the training process by reducing the wall-clock time of iterative local training of all the participants. The acceleration results are shown in Fig.\ref{fig5}. The left part in Fig.\ref{fig5} shows that increasing the runner-level parallelism can well reduce the time cost as the number of runners grows. The right part in Fig.\ref{fig5} shows that increasing the number of processes can also save the training efficiency, especially when the cost of data/model transmission across different devices is relatively smaller than the local training processes.
\begin{table*}
\centering
\begin{tblr}{
  cell{1}{2} = {c=3}{c},
  cell{1}{5} = {c=3}{c},
  cell{1}{8} = {c=3}{c},
  vlines,
  hline{1-3,6} = {-}{},
}
    & \textbf{Cifar10 - CNN} &                    &                   & \textbf{Cora - GCN} &                    &                & \textbf{Shakespeare - LSTM} &                    &                \\
\textbf{Data Heterogeneity} & \textbf{IID}                   & \textbf{Imbalance} & \textbf{Dir(0.3)} & \textbf{IID}        & \textbf{Imbalance} & \textbf{Louv.} & \textbf{IID}                & \textbf{Imbalance} & \textbf{Div.} \\
\textbf{fedavg}             & 0.8216                               &  0.8303                  &  0.7355                 &   0.9059                  & 0.9022                   &   \textbf{0.7693}             &      \textbf{0.4457}                       &  0.4424                  &  \textbf{0.3340}              \\
\textbf{fedprox}            &  0.8206                          & 0.8207                   &  0.7099                 &   0.8911                  &  0.9022                  &   0.7638             &  0.4443                 & \textbf{0.4432}                   &  0.2934              \\
\textbf{scaffold}           &   \textbf{0.8551}     & \textbf{0.8474}                   &  \textbf{0.7492}                 &  \textbf{0.9151}                   &   0.9022                 &    0.7638            &  0.4304                   &  0.4309                 &  0.2775              
\end{tblr}
\caption{Comparison on model testing accuracy of different methods across datasets and heterogeneity settings.}
\label{table5}
\end{table*}
\subsubsection{Experiment Tracker and Analyzer}
We realize two modules to help do experiments. First, we use the logger to track the running-time variables of interest without writing intrusive codes, and we preserve APIs to customize loggers for different experimental purposes. The logger module is also compatible with popular ML logging tools like wandb \cite{wandb}. The running-time variables of interest will finally be stored on the disk as records. Second, we develop the analyzer to load the experimental records from the disk and help make further analyses on them. We also provide basic tools to visualize the records and generate tables in a customizable manner.
\subsubsection{Parameters Auto-Tuning}
We realize automatically tuning hyper-parameters based on the aforementioned parallelism. By using this module, developers can input the grid of hyper-parameters and specify the GPUs to find the group of hyper-parameters that achieves the optimal performance on the validation dataset. 

\section{Experiments}
In this section, we conduct experiments to show how FLGo facilitates FL studies by customization from three aspects: data heterogeneity, system heterogeneity, and asynchronism. The meanings of heterogeneity notions are in the appendix.
\subsection{Experimental Settings}
\subsubsection{Datasets and Models}
The datasets and corresponding models are concluded in Tabel.\ref{table4}.
For data heterogeneity experiments, we respectively conduct I.I.D, quantity skew, and non-I.I.D scenarios for three datasets. For CIFAR10, we use I.I.D, Imbalance (0.5), and Dirichlet (0.3) data partition. For Cora, we employ IID, Imbalance, and Louvain data partition. For the Shakespeare dataset, we use IID, Imbalance, and Diversity (0.2) data partition. 
For system heterogeneity and asynchronism experiments, we use I.I.D and Dirichlet (1.0) data partition.

\subsubsection{Baselines}
For data heterogeneity experiments, the baselines are FedAvg\cite{mcmahan2017communication}, FedProx\cite{li2020federated}, and Scaffold\cite{karimireddy2020scaffold}. For system heterogeneity experiments, the baselines are FedAvg\cite{mcmahan2017communication}, FedNova\cite{wang2020tackling}, and FedProx\cite{li2020federated}. For asynchronism experiments, the baselines are synchronous algorithms FedAvg\cite{mcmahan2017communication}, FedProx\cite{li2020federated} and the asynchronous algorithm FedAsync\cite{Xie2019AsynchronousFO}. 

\subsubsection{Hyparameters}
We tune all the methods by grid search. The hyperparameters for the CIFAR10 and Cora datasets: the number of rounds to be 1000, the learning rate to be 0.1, proportion to be 1.0,epoch $E \in$$\left\{1,3,5\right\}$, and batch size to be 50. For the Shakespeare dataset, we set rounds to 100, learning rate $\eta \in$$\left\{0.1,0.3,0.8\right\}$, proportion to be 0.1, epoch $E\in$$\left\{5,10\right\}$, and batch size to be 64. For Fashion Mnist and SVHN dataset, the number of rounds to be 1000, the learning rate to be 0.1, epoch to be 5, and batch size to be 50. For Reddit dataset, the number of rounds to be 1000, the learning rate, epoch and batch size to be 1, 1 and 50, respectively. 
\subsubsection{Implementation} Experiments are run on a 64 GB-RAM Ubuntu 18.04.6 server with Intel(R) Xeon(R) CPU E5-2630 v4 @ 2.20GHz, 4 NVidia(R) 2080Ti GPUs, and PyTorch 1.12.0. 
\subsection{Data Heterogeneity}
We compare the performance of three state-of-the-art FL algorithms toward the non-IID problem in Table.\ref{table5}. 
The algorithm's performance varies significantly with different datasets and distribution settings. For CIFAR10, Scaffold achieves the best performance under all the data distributions. For Cora, all the methods achieve similar model performance. Scaffold achieves the best results on the IID distribution, while FedAvg performs the best on the Louvain distribution. For Shakespeare, the best results lie in FedAvg and FedProx, where Scaffold is the worst one across all the data distributions. These results suggest that there is no method that always outperforms other methods across datasets and heterogeneity settings, which is consistent with the observations in \cite{li2022niidbench}. In addition, the imbalance setting won't have an obvious impact on the results of all these methods. The non-IID settings cause performance reduction for all the methods on different datasets compared to the IID settings.

\subsection{System Heterogeneity}
\begin{figure*}[htbp]
\centering
    \subfigure[The impact of client availability]{
    \centering
    \includegraphics[width = .24\textwidth]{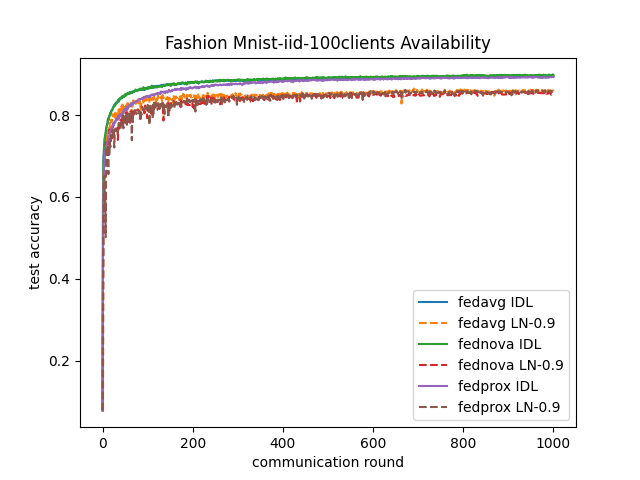}
    \includegraphics[width = .24\textwidth]{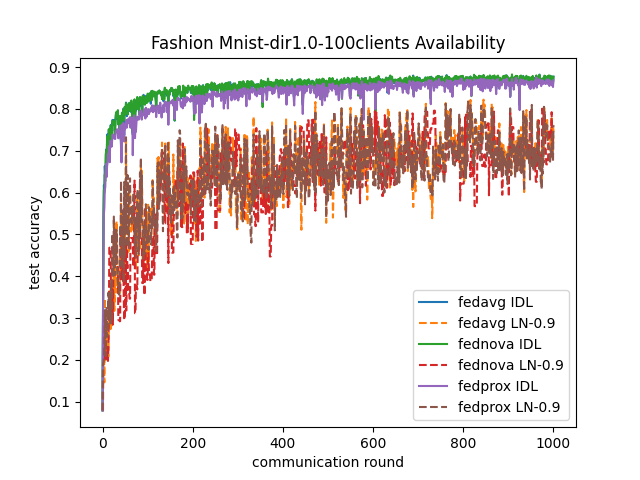}
    \includegraphics[width = .24\textwidth]{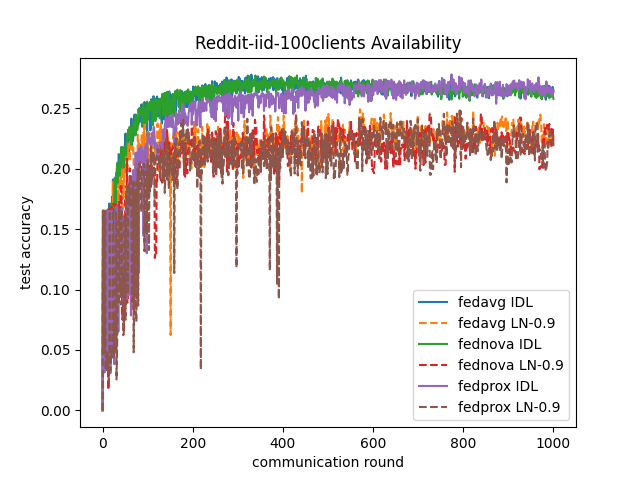}
    \includegraphics[width = .24\textwidth]{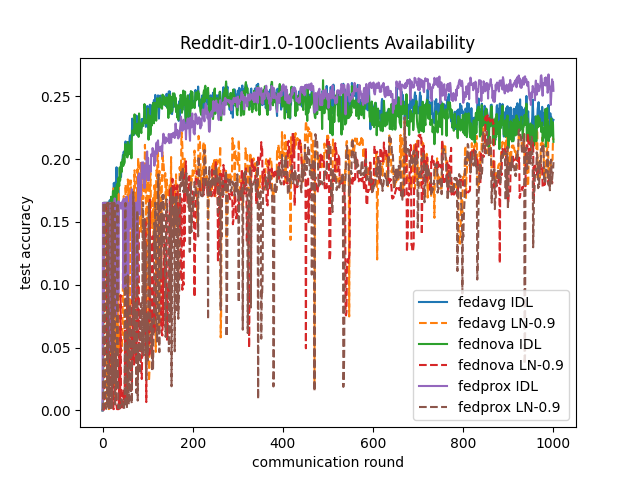}
    
    \label{fig6[a]}
    }
    \subfigure[The impact of client completeness]{
    \centering
    \includegraphics[width = .24\textwidth]{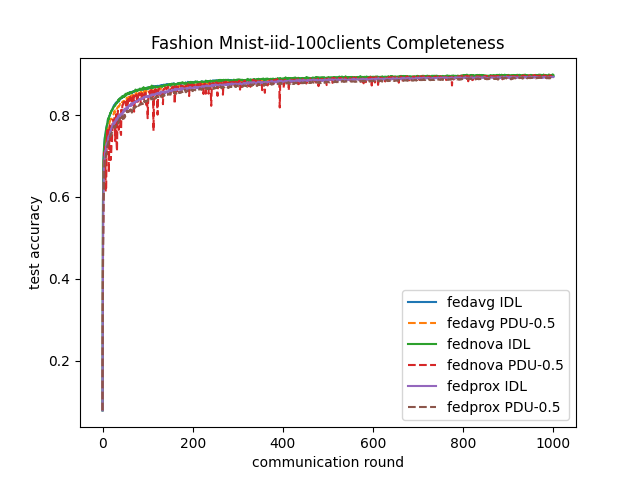}
    \includegraphics[width = .24\textwidth]{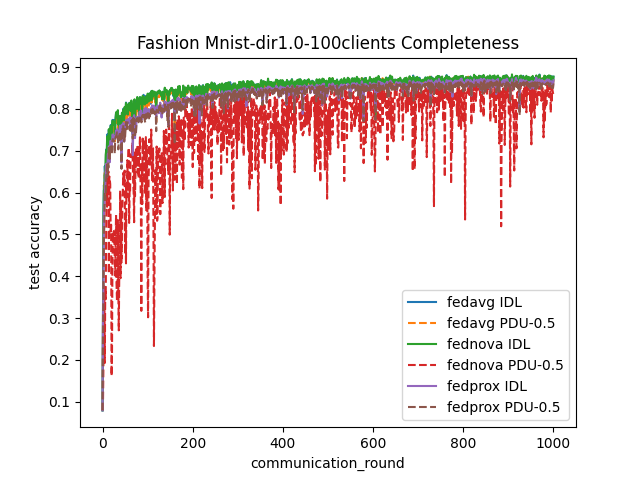}
    \includegraphics[width = .24\textwidth]{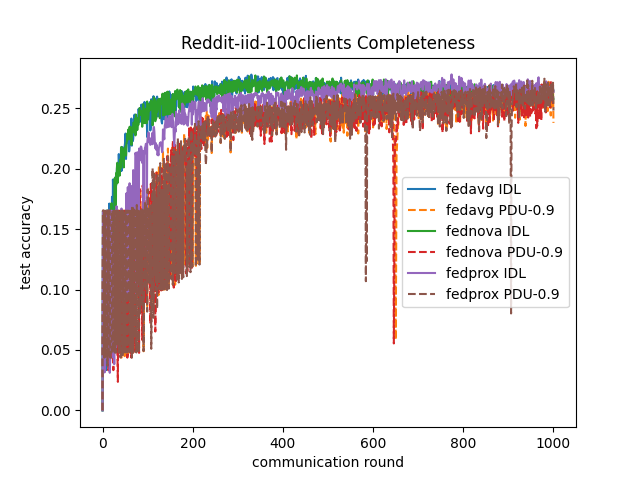}
    \includegraphics[width = .24\textwidth]{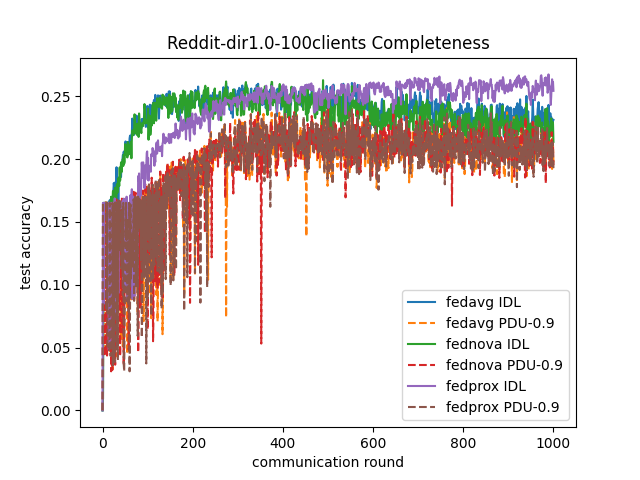}
    \label{fig6[b]}
    }
    \subfigure[The impact of client connectivity]{
    \centering
    \includegraphics[width = .24\textwidth]{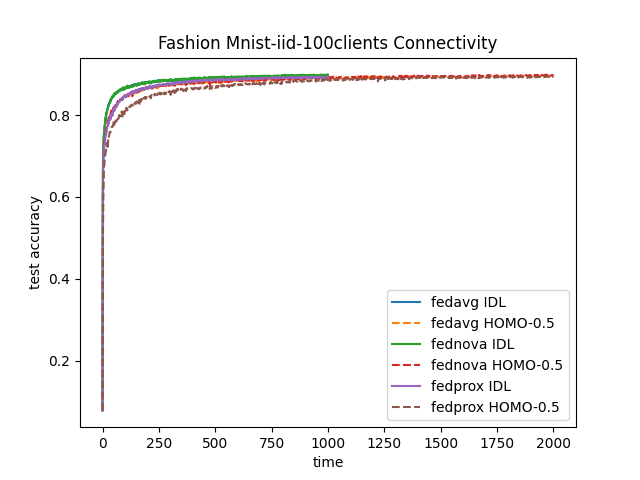}
    \includegraphics[width = .24\textwidth]{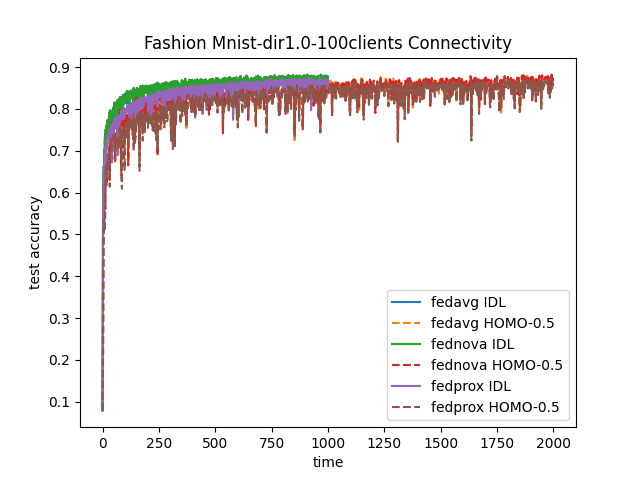}
    \includegraphics[width = .24\textwidth]{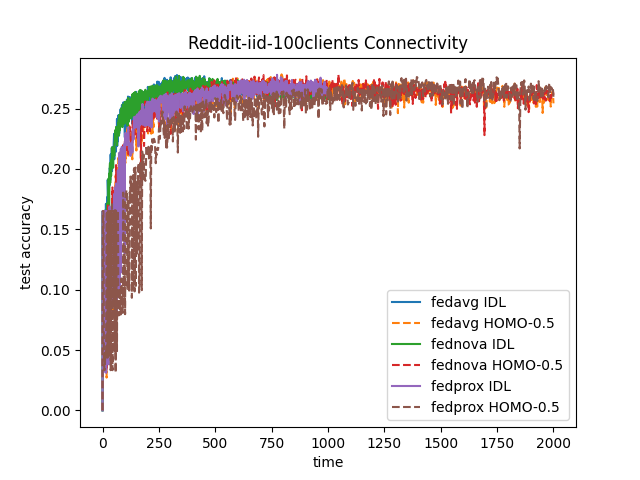}
    \includegraphics[width = .24\textwidth]{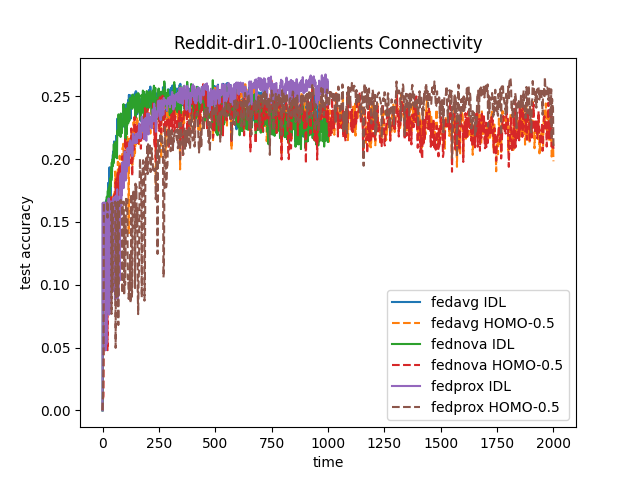}
    \label{fig6[c]}
    }
    \subfigure[The impact of client responsiveness]{
    \centering
    \includegraphics[width = .24\textwidth]{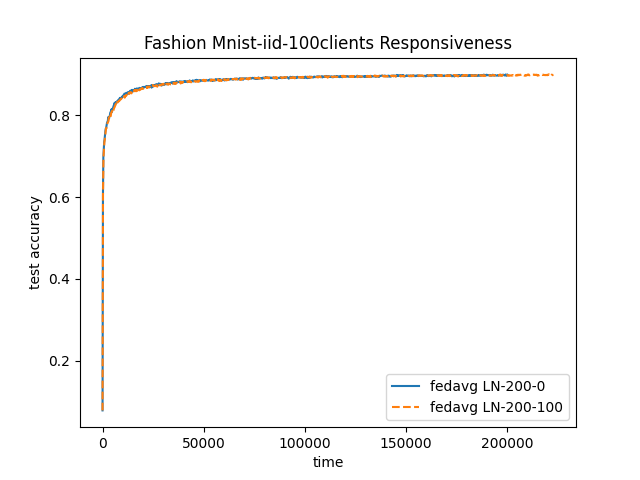}
    \includegraphics[width = .24\textwidth]{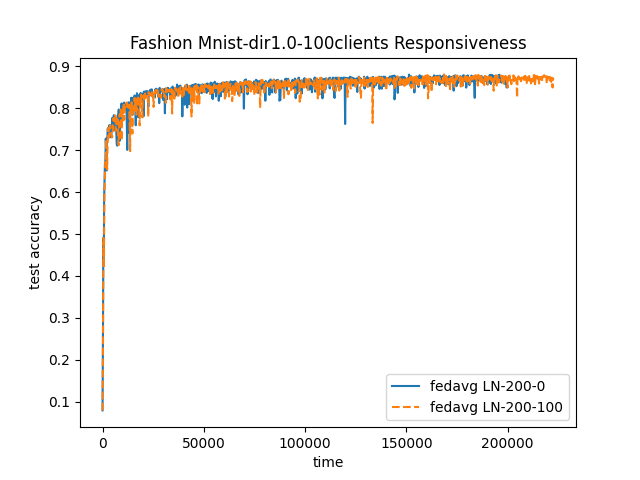}
    \includegraphics[width = .24\textwidth]{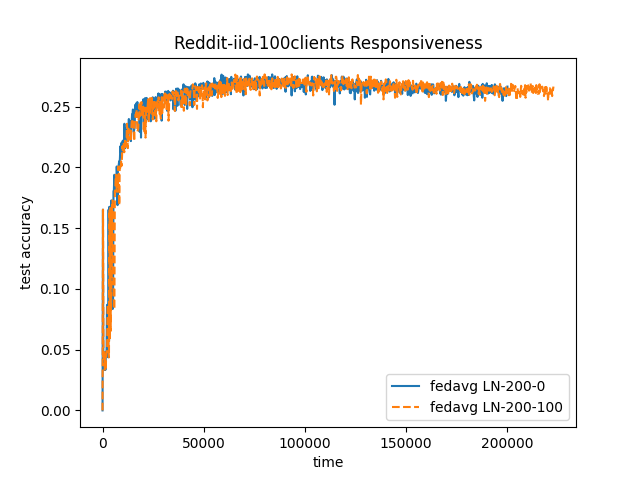}
    \includegraphics[width = .24\textwidth]{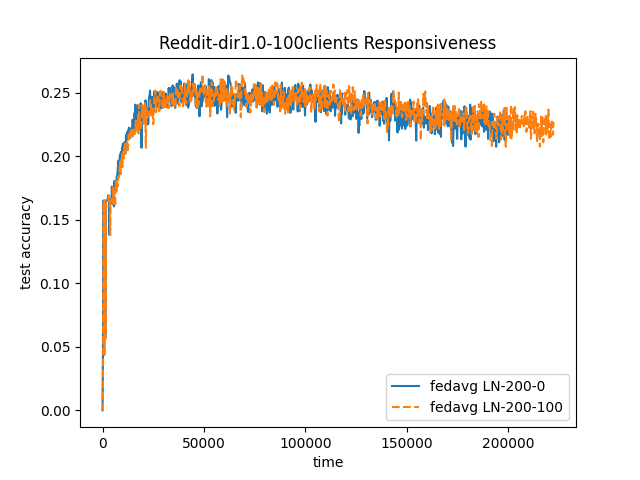}
    \label{fig6[d]}
    }
    \subfigure[Asynchronism]{
    \centering
    \includegraphics[width = .24\textwidth]{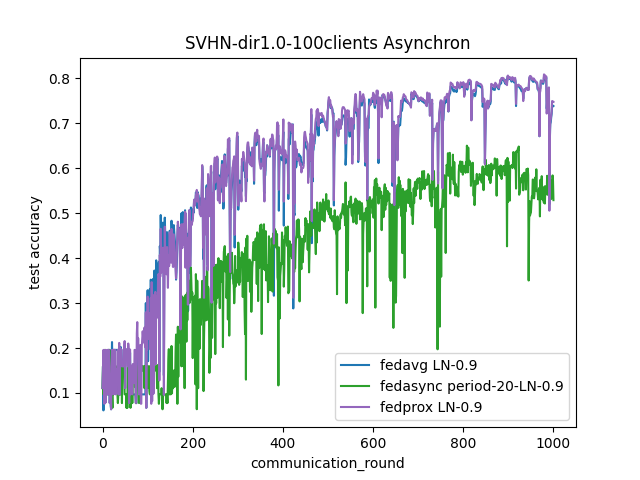}
    \includegraphics[width = .24\textwidth]{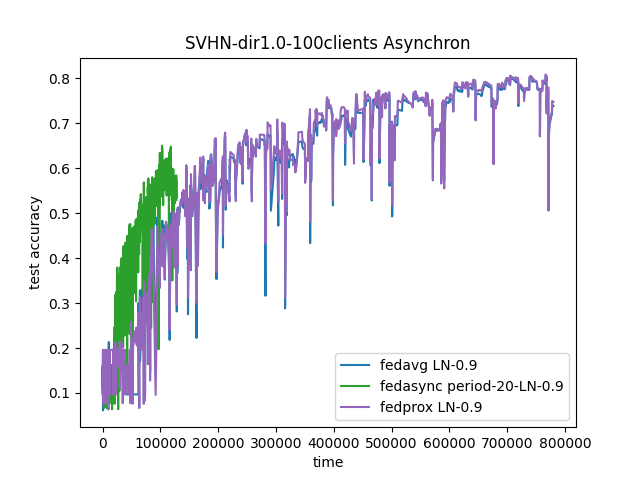}
    \includegraphics[width = .24\textwidth]{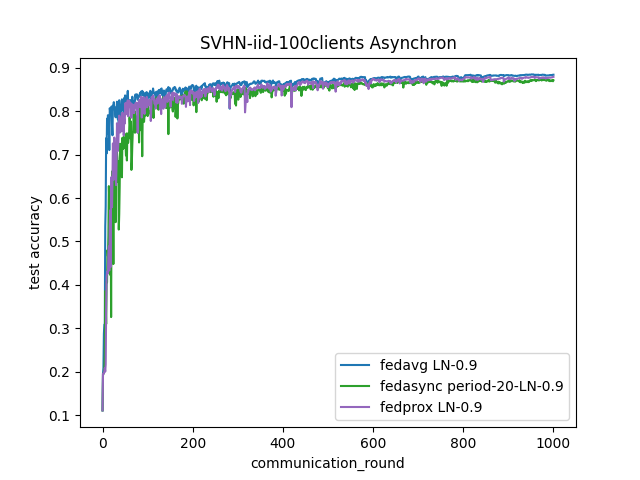}
    \includegraphics[width = .24\textwidth]{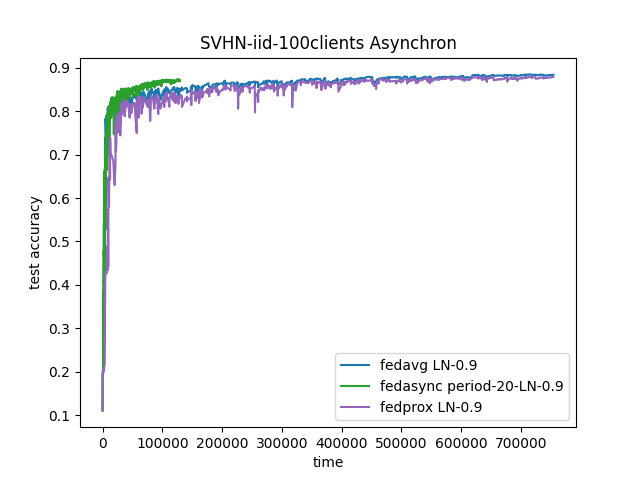}
    \label{fig6[e]}
    }
    \label{fig6}
    \caption{The subfigures in the first 4 rows conduct the impact of system heterogeneity, where each row's heterogeneity is of: (a) availability; (b) completeness; (c) connectivity; and (d) responsiveness. For each row, in the order from left to right: IID-Fashion, Dirichlet-Fashion, IID-Reddit, and Dirichlet-Reddit. (e) compares synchronous and asynchronous strategies in the simulation of the complex combinations of different heterogeneity.}
\end{figure*}

We respectively investigate the impact of 4 different types of system heterogeneity on three FL algorithms in FashionMNIST and Reddit datasets: FedAvg\cite{mcmahan2017communication}, FedNova\cite{wang2020tackling} and FedProx\cite{li2020federated}.
Fig.\ref{fig6[a]} suggests that the heterogeneity of client availability has a non-trivial negative impact on the performance of all the methods. Especially, when data heterogeneity meets availability heterogeneity, the performance is further worsened than the IID scenario on both two datasets. 
Fig.\ref{fig6[b]} shows the impact of heterogeneity in client completeness. When clients are not able to complete all the local training epochs, the training efficiency is correspondingly degraded. Although FedNova is claimed to be robust to varying completeness, its performance is slightly worse than FedAvg and FedProx in FashionMNIST, which may be due to the over-enlarging of short model updates.
Fig.\ref{fig6[c]} shows the impact of heterogeneity in client connectivity. We use the virtual time as the horizontal axis instead of the communication round since the server will wait for another unit of time once there exist dropping-out clients. The results show that less connectivity will increase the time cost of the training process. For FashionMNIST, the connectivity has only a trivial impact on the model performance. For Reddit, less connectivity causes performance degradation in the non-IID setting.
Fig.\ref{fig6[d]} shows the impact of heterogeneity in client responsiveness. This heterogeneity will only influence the time cost of training since the server will wait for all the clients. The results show that a larger variance of client latencies leads to larger time costs with the same client latency mean. This is consistent with the \textit{Cask Effect}
 where the slowest client will dominate the time costs for synchronous algorithms.
\subsection{Asynchronism}
We create a complex combination of different heterogeneity to customize the environment. First, we make the probability of each client follow the distribution $lognormal(0, -log(0.1))$.  We use the same completeness heterogeneity setting in \cite{li2020federated}. We make the clients have the same drop probability of $0.5$, and we set the response latencies of clients to follow a log-normal distribution with $(mean, var.)=(200, 50)$. The results in Fig.\ref{fig6[e]} show that asynchronous strategies can significantly reduce time costs when carrying out the same times of aggregation. Under the IID setting, FedAsync reduces the time cost to achieve the same model performance against other baselines. Under the non-IID setting, the performance of FedAsync is worse than FedAvg and FedProx after the same amounts of aggregations, which suggests that the effectiveness of aggregation for asynchronous strategies can be further improved.  

\section{Conclusion}
In this work, we presented a novel lightweight FL platform, FLGo, to facilitate cross-application FL studies with a high ability of shareability . FLGo offers 40+ benchmarks, 20+ algorithms, and 2 system simulators as out-of-the-box plugins. We also developed a range of experimental tools for various experiment purposes. Comprehensive experiments were conducted to verify the ability of customization of FLGo, under various data and system heterogeneity settings.
\appendix


\section*{Acknowledgments}
The research was supported by Natural Science Foundation of China (62272403, 61872306).


\bibliographystyle{named}
\bibliography{ijcai23}

\end{document}